\documentclass[11pt,a4paper]{article}
\usepackage[hyperref]{naaclhlt2018}
\usepackage{times}
\usepackage{latexsym}
\usepackage{graphicx}
\usepackage{url}

\aclfinalcopy 

\newcommand\NemexRelator{{\sc LightRel}}
\long\def\comment#1{}

\title{\NemexRelator{} at SemEval-2018 Task 7: Lightweight and Fast Relation Classification }

\author{Tyler Renslow \\
  DFKI, Saarbr\"ucken, Germany\\
  {\tt tdrenslow@gmail.com} \\\And
  G\"unter Neumann \\
  DFKI, Saarbr\"ucken, Germany\\
  {\tt neumann@dfki.de} \\}

\date{}

\begin{document}
\maketitle
\begin{abstract}
We present \NemexRelator{}, a lightweight and fast relation classifier. Our goal is to develop a high baseline for different relation extraction tasks. By defining only very few data-internal, word-level features and external knowledge sources in the form of word clusters and word embeddings, we train a fast and simple linear classifier.

\end{abstract}

\section{Introduction}

The main motivation for our participation at SemEval-2018 \cite{SemEval2018Task7} was the ideal opportunity to test and improve our relation extraction system \NemexRelator{}. The system design and development was inspired by the work described in \cite{NguyenG15}. Their goal was to depart from traditional relation extraction approaches with complicated feature engineering by exploring a deep neural network that would minimize its dependence on external toolkits and resources, e.g. external word embeddings. That allowed them to design a rather lightweight relation extraction approach that would basically only require supervised training data, external word embeddings and a few hyperparameters. Their "end-to-end" relation extraction approach produced competitive results.

Since tuning hyperparameters for neural networks can be a intricate process, we considered whether it would be possible to define an even simpler system and use it as a baseline for our future research. Thus, we adopted some of the design decisions made by \cite{NguyenG15} and combined them with a well-known, fast linear classifier, viz. LibLinear \citep{Fan2008}. 

Following \newcite{NguyenG15}, we represent a relation mention as a sequence of tokens.
The core idea of our approach consists of transforming this sequence into a a vector of fixed length, such that each token (or word) is represented only by: 1) the word itself, 2) its shape (a small, fixed amount of character-based features), 3) the word's cluster id, and 4) the word's embedding of fixed size.

For this competition, we introduce a new relation-level feature, namely the ID of the word directly following and preceding entities one and two, respectively.
Furthermore, we ignore all tokens to the left of the first entity and to the right of second entity.
The size of the whole vector therefore hinges on the maximum number of elements between the two entities found in the training set.

These representations are then used to train a LibLinear model. 
Note that this reduces manual feature engineering to defining the shape features, finding an appropriate number of clusters and word embedding dimensions, and hyperparameters for LibLinear.
All other information is automatically computed from the training data. In this sense, we consider our system lightweight.

We initially developed and tested our approach on the previous and widely-used SemEval-2010 Task 8 data set \cite{HendrickxKKNSPP10}, and obtained as our best result an F1 measure of 79.78\% on the test-data using the standard evaluation script from SemEval-2010 (see also Sec. \ref{sec:experiments}). 
Although this result is behind the best reported ones (the majority between 83\%-85\%, and the best 88.0\%, cf. \cite{WangCML16}), we think it provides a strong baseline compared to the manually-engineered, feature-heavy approaches or complex neural architectures.
Thus, when the SemEval-2018 Task 7 challenge was announced, it was a natural decision to use it as an additional testing ground for \NemexRelator{}.

\section{Approach}

\NemexRelator{} can be divided into three major steps:
\comment{
\begin{enumerate}
    \item extracting and storing information from the training data and external sources in an internal representation;
    \item converting the internal representation into feature vectors;
    \item training a logistic regression classification model to predict classes using LibLinear
\end{enumerate}
}
1) extracting information from the training data and external sources and storing it in an internal representation; 
2) converting the internal representation into feature vectors;
3) using feature vectors to train a logistic regression classification model to predict classes.

In the first step of the system, pertinent information is extracted from the training data.
Each relation instance (the two entities and the text between them) is collected.
Along with the relation instance, the ID of the abstract, the IDs of the entities, the relation type and the length of the sentence are all gathered into one internal representation to facilitate vector computation later.
We do not include additional data in our training set, e.g. from other tasks or subtasks, previous SemEval years, etc.
For example, the following relation:

{\small
\begin{verbatim}
<entity id="E89-1006.1">French tenses
</entity> in the framework of <entity 
id="E89-1006.2">Discourse 
Representation Theory</entity>
\end{verbatim}
}

\noindent would be represented in our system as:

{\small
\begin{verbatim}
('E89-1006', ['French_tenses', 'in', 
'the', 'framework', 'of', 
'Discourse_Representation_Theory'], 
'E89-1006.1', 'E89-1006.2', 
'MODEL-FEATURE REVERSE', 6)
\end{verbatim}
}

The only processing done on the text is: 1) merging any punctuation and the word before it into a single string and 2) joining multi-word entities into a single string with an underscore.  
Using the words from these instances, we index a unique vocabulary and the single word immediately following or preceding entity one or entity two, respectively, provided that the word isn't the other entity. 
The unique relation types are also indexed later so that their unique identifiers can be used as a feature in the training vectors for LibLinear.
In the case of the competition, the test data is used to expand the unique vocabulary and entity context words. 
Once this information is collected, it can be converted into vector representations to train a LibLinear model.

The next step involves converting our relation instances into feature vectors.
In addition to the information gleaned from the training data, we use features that are independent of our training data.
For instance, we include a word-shape feature, a unique vector representing certain character-level features found in a word.
In particular the features are based on whether: any character is capitalized; a comma is present; the first character is capitalized and the word is the first in the relation (representing the beginning of a sentence); the first character is lower-case; there is an underscore present (representing a multi-word entity); and if quotes are present in the token.
These features were left unchanged from the ones that achieved the best results on SemEval-2010 Task 8.

We also incorporate our own word embeddings into our feature vectors.
Our previous system developed for the SemEval-2010 task used Numberbatch embeddings from ConceptNet \cite{speer2017conceptnet} to yield the best results, but development was slower than desired due to the size of the embedding file.
We therefore hand-pick the data used to calculate the new, smaller embeddings in order to experiment with domain-specific word embeddings better tuned to the task at hand.

We pre-compute two word embedding files: one based on the ACM-Citation-network V9 corpus of abstracts and the other on the DBLP-Citation-network V5 corpus.\footnote{Corpora available at \url{https://aminer.org/citation}.}
The embeddings are calculated using the word2vec\footnote{\url{https://code.google.com/archive/p/word2vec/}} tool with the following constraints: the continuous bag-of-words model, 300-dimension vectors (chosen for portability from the old system that used 300-d Numberbatch vectors) and leaving out tokens occurring fewer than five times.
The DBLP corpus embedding file used for the final system was around half the size of the Numberbatch embeddings.

Cluster-membership features are also included in our feature vectors.
We used the MarLin \cite{muller-schuetze:2015:NAACL-HLT} clustering tool to pre-compute word clusters for the aforementioned two corpora based on their bigram context.
In particular, we ran five training epochs to cluster the words into 1000 classes.
MarLin was chosen over the Brown clustering algorithm \cite{Brown1992}, as these clusters produced better results for SemEval-2010 Task 8.
Using all of our features, we convert our internal representation into the proper input format for training a LibLinear model.
All features are binary besides the word embeddings. The test data is converted into the same format for predicting, albeit without relation IDs in the case of the competition run.

In the same way as \cite{NguyenG15}, we represent a relation mention $x$ of length $n$ as a sequence of tokens $x = [x_1, x_2, ..., x_n]$, where $x_i$ is the $i$-th word in the mention.
Furthermore, let $x_{i{_1}}$ and $x_{i{_2}}$ be the head of the two entity mentions of interest.
Now, we transform this into a vector of fixed length $l$, whose size is determined by the relation mention with highest number of tokens $k$ between its entity head tokens, and by using $L$ tokens left to the entity token $x_{i{_1}}$, and $R$ tokens to the right of $x_{i{_2}}$.
In all experiments described below, $L$ and $R$ are set to 0, which is also done by \cite{NguyenG15}.
If a relation mention has fewer than $k$ elements between entities, we add padding elements (i.e. dummy word tokens).
One of the system parameters we tuned was the padding strategy employed: in the end, we got better results when padding after entity one as opposed to padding before entity two.
Thus, initially, all relation mentions are represented by a fixed length vector $x' = [x'_1, x'_2, ..., x'_{k+2}]$, where $x'_1 = x_{i{_1}}$ and $x'_{k+2} = x_{i{_2}}$, with words plus the necessary number of padding elements between.

Once we have a fixed-length vector for each relation, we append the word-context feature to the vector.
The motivation behind this feature is to provide some distributional information to the model regarding the syntactic contexts in which entities occur.
In the case where there are no words between entities, we use no word context in our feature vector.
If a single word occurs between entities, they share the same context.
Otherwise, word context is represented by the word directly to the right and left of entity one and two, respectively.
The context features are calculated based on the original, non-normalized sentence, so that padding elements are not involved.

The last step involves training a model to predict classes based on our vector representations of the training data.
We employ LibLinear's default classifier (version 2.11), cf. also Sec. \ref{sec:experiments}.
Once the model is trained, it can then make predictions on the vectors that represent test relation instances.
The predictions are then converted back into the format necessary for evaluating our system using the scorer script provided by SemEval.

\section{Experiments}
\label{sec:experiments}

As mentioned before, our system was an adaptation to the one that produced the optimal result on SemEval-2010 Task 8.
Through cross-validation, we tuned our system parameters to produce the best results on SemEval-2018 training data, bearing in mind the goal of keeping our system as lightweight as possible.
It is important to note that hyperparameter tuning was only done according to system performance on task 1.1's training data; we simply used the optimal parameters from this task while participating in task 1.2 for the competition.

Through experimentation, we found a LibLinear classifier that performed better on the current task than the one that we used for the SemEval-2010 task.
We also developed a novel word-context feature for the 2018 task, which modestly improved our cross-validation results (anywhere from a 1\% to 3\% increase in F1 score, depending on the different parameters used).
We will display the parameters that produced optimal results in cross-validation, as well as the results obtained using these same parameters in the competition phase in tabular form below.

For the SemEval-2010 task, we obtained the best results using LibLinear's support vector classifier by Crammer and Singer \cite{DBLP:conf/colt/CrammerS00} with a cost of 0.1 and a stopping tolerance of 0.3.
Given the fewer relation types and smaller training vectors in the SemEval-2018 task, we experimented with different classifiers. 
In development, we achieved the best performance in using LibLinear's default classifier, which performs dual L2-regularized L2-loss support vector classification, cf. \cite{Fan2008}. 
We set cost and stopping tolerance parameters equal to 0.1, using default settings for all other parameters.

The total competition data was made up of $1228+355=1583$ relation instances (\#training + \#test), corresponding to an approximate train-test split of $\approx$78\%-22\%.
Because of this, we developed our system using 5-fold cross-validation on the training set, which entails an even 80\%-20\% split of data, i.e. $982+246=1228$ instances.
Even though a subset of the training data was provided by the organizers for system development, we opted to split the data in accordance with the proportion of training and test instances, in order to get the best estimate of system performance in the competition phase.

We obtained the best average F1 score for tasks 1.1 and 1.2 when using all features but the shape feature (word, embeddings, clusters, entity one context, entity two context); the second best score was obtained when using all features and the third best by removing the entity two context feature from the entire feature set.
These feature sets represent a modest approach to the task at hand compared to more complex systems incorporating knowledge from sources like part-of-speech tagging or dependency parsing.
An exception to these feature sets was cross-validation on task 1.2's training data, where an average F1 score of 61.83\% was obtained when using neither context-related feature.
However, since removing a context-related feature (no e2 context) already produced the best results in development, we decided to hedge our bets in the competition with a feature set composed of all features but the shape feature, i.e. of no manually engineered features. 

The results from cross-validation on task 1.1 and task 1.2 are shown in Table \ref{tab:cv_res} below.
Based on these results, we expected our system to perform similarly in the competition.

\begin{table}[h!]
\centering
\begin{tabular} { |c|c|c| }
    \hline
    feature set & task 1.1 & task 1.2 \\
    \hline
    all features & 45.4\% & 62.0\%\\
    w/o shape & 46.4\% & 61.7\%\\
    w/o e2 context & 45.5\% & 62.1\%\\ 
    \hline
\end{tabular}
\caption{Average F1 scores from 5-fold cross-validation.}
\label{tab:cv_res}
\end{table}

The actual results of the competition can be seen below in Table \ref{tab:comp_res}.
These results placed us in 18th out of 28 groups in subtask 1.1 and 12th out of 20 in subtask 1.2.

\begin{table}[h!]
\centering
\begin{tabular} { |c|c|c| }
    \hline
    feature set & task 1.1 & task 1.2 \\
    \hline
    all features & 39.3\% & 67.5\%\\
    w/o shape & 39.9\% & 68.2\%\\
    w/o e2 context & 39.2\% & 67.5\%\\ 
    \hline
\end{tabular}
\caption{Competition F1 scores.}
\label{tab:comp_res}
\end{table}

The best result was obtained when using all features except the shape feature.
This points to evidence that there is overlap in the information gained from the shape feature and the word feature.
The same token, differing only in punctuation (e.g. the strings 'IR' and 'IR,'), is represented with both different word and different shape IDs in our system.
However, for the shape feature to provide extra information to the model, the word feature would have to remain the same, since the shape feature changed.

The results on the first task were worse than the cross-validation results suggested.
Since we incorporated the words from the test data into our word and context features, there was no information that the model could have missed in the competition phase.
Therefore, we attribute the slight decrease in performance to the fact that more training and less test data were used in development, meaning that our models were overfitting in training.

Surprisingly, our system performed better on noisily annotated data, given no extra development in relation to the task.
It is difficult to say with conviction why these results occurred, as our features do not incorporate the entity markup.
Another difference in this task is the data itself; it could be that more tokens were found in the embedding and cluster features, providing more information to the model.
However, this fact alone hardly explains an almost 30\% increase in F1.


Finally, we assessed our system's speed.
The final system which produced the best results for subtask 1.1 needed a total of 35 seconds to run on a 2012 MacBook Pro with 16GB of RAM and a 2.6 GHz quad-core Intel Core i7 processor\footnote{implementation available at \url{https://github.com/trenslow/LightRel}}.
The bottleneck occurred in the creating of vectors (80\% of total time), which can be attributed to the simple way we stored and accessed the embeddings.
Training lasted 5 seconds, while testing/prediction only required a fraction of a second.
These results demonstrate our system's agility in relation to complex neural architectures, which typically need hours, or even days, to train.

\section{Conclusion}
We believe our system has established a useful baseline for relation classification. 
Our approach is simple in that it involves few features.
These few features yield remarkable results given the amount of time required to deploy the system, allowing for quicker development and prototyping of models compared to more cumbersome neural networks.
The performance on task 1.2 as opposed to task 1.1 demonstrates our system's flexibility, as we obtained fair results with no extra development.
However, further research is needed to explain the jump in performance between the tasks.

\section*{Acknowledgments} 
This work was partially funded by 
the BMBF through the project DEEPLEE (01IW17001) and the European Union's Horizon
2020 grant agreement No. 731724 (iREAD).

\bibliographystyle{acl_natbib}

\end{document}